# A Pilot Evaluation of ChatGPT and DALL-E 2 on Decision Making and Spatial Reasoning

Zhisheng Tang[1], Mayank Kejriwal[2]

University of Southern California

## Abstract

We conduct a pilot study selectively evaluating the cognitive abilities (decision making and spatial reasoning) of two recently released generative transformer models, ChatGPT and DALL-E 2. Input prompts were constructed following neutral *a priori* guidelines, rather than adversarial intent. *Post hoc* qualitative analysis of the outputs shows that DALL-E 2 is able to generate at least one correct image for each spatial reasoning prompt, but most images generated are incorrect (even though the model seems to have a clear understanding of the objects mentioned in the prompt). Similarly, in evaluating ChatGPT on the rationality axioms developed under the classical Von Neumann-Morgenstern utility theorem, we find that, although it demonstrates some level of rational decision-making, many of its decisions violate at least one of the axioms even under reasonable constructions of preferences, bets, and decision-making prompts. ChatGPT's outputs on such problems generally tended to be unpredictable: even as it made irrational decisions (or employed an incorrect reasoning process) for some simpler decision-making problems, it was able to draw correct conclusions for more complex bet structures. We briefly comment on the nuances and challenges involved in scaling up such a 'cognitive' evaluation or conducting it with a closed set of answer keys ('ground truth'), given that these models are inherently generative and open-ended in responding to prompts.

[1] zhisheng@isi.edu
[2] kejriwal@isi.edu

1. **Background**

The development and release of the attention-based transformer neural network architecture in 2017 has since led to a stunning explosion of such models [1]. An early popular example is the Bidirectional Encoder Representations from Transformer (BERT) model [2], which soon led to many domain-specific variants, as well as a more optimized version that was able to yield significant improvements without major changes to the original BERT architecture [3]. Perhaps because of its success, researchers have been attempting to empirically understand the properties (including biases and blind spots [4]) of even early transformer models such as BERT, along multiple dimensions [5-7]. While these tests, some of which have been adversarial by design, have revealed some problems, a growing body of research also shows that these models have achieved truly impressive, non-incremental performance advances on various natural language understanding problems [8].

While it can be convenient to overweight mistakes by the models, especially if the mistakes are 'un-humanlike' and made in seemingly simple situations, and to dismiss them as incapable of semantics or symbolic processing, such commentating potentially opens the door to confirmation bias. We are not denying the utility of critical and adversarial testing of such models [9,10]; however, we do caution that there is a danger of their interpretations being taken out of context. Arguably, the latest transformer models, such as ChatGPT and DALL-E, captured the public spotlight by being able to process relatively complex human inputs with unprecedented skill [11]. They have also ignited an AI arms race of sorts between large technology corporations. Some of this discourse is hyped, but some could be argued to be justified as correctly describing a major leap in AI progress, at least in an empirical sense [12, 13]. On the academic front, large language models have garnered interest well beyond industry, education, and application. In recent international conferences on cognitive science [14], for instance, they have been given their own sessions, and at least a few papers and abstracts from that community have attempted to study them from a cognitive perspective e.g., [15,16].

With this emergence in the literature in mind, we propose that such models should rightfully be evaluated as 'cognitive machines' i.e., rather than probing the models in an ad-hoc fashion, we instead propose the use of systematic *cognitive* tests without *a priori* knowledge of how the model will perform, inspired by a long history of similarly evaluative work in both animal cognition and human psychology, often with a utility in mind, such as early detection of Alzheimer's or spaceflight readiness [17,18,19]. We draw a distinction at the outset of claiming that such models *are* cognitive machines. Rather, we merely claim that they should be evaluated as such precisely because the extent of their ability to reason in a humanlike manner continues to be controversial and subject to the investigator and confirmation bias.

We note that systematic benchmarking of such models on applied AI tasks (e.g., question answering, text summarization) is already the norm in computer science, but has come under criticism itself for general AI problems like commonsense reasoning [20]. Unfortunately, a concordant degree of discipline has not been applied in studying these models *qua* cognitive machines. Rare examples of such systematic testing include[3] [21,22]. Another example of comparing different model outputs is [23]. Others have attempted such studies for very specific problems like multiple-choice question answering [24,25].

---

[3] Other very systematic examples include [23], which compared different models' outputs. Yet other attempts are for studies in very specific modalities like multiple-choice question answering [24,25].

In this preliminary study, we conduct a pilot evaluation by constructing and applying two cognitive tests (spatial reasoning and decision making), one of which is more appropriate for a text-to-image generative model such as DALL-E 2, with the other being more appropriate for models that can better express logic and abstraction in common language. Both tests require text as input, but we hypothesize that spatial reasoning is more directly tested through the production of visual output, while decision-making is better tested through contextualized conversation-style text output and is hence more appropriate for a large language model like ChatGPT:

- *Spatial reasoning:* This test is comprised of a set of pairs of prompts, where each prompt expresses a spatial relationship between common objects, such as apples or oranges. The pair of prompts use the same objects and is largely indistinguishable in much of their surface prompts, but contains a single contrast e.g., a top relation versus a bottom relation between objects. We qualitatively assess both the extent to which DALL-2 is able to understand the prompt in isolation and the difference in outputs between the two prompts, thereby aiming to test if the model is able to convincingly differentiate between the relations.
- *Decision-making:* This test uses a dialectic sequence of prompts to understand whether the model obeys some of the axioms of Von-Neumann rationality (VNM), first elucidated under the Von Neumann-Morgenstern utility theorem [26]. A detailed description of these axioms is beyond the scope of this paper, but we briefly describe the intuition behind the axioms we do test when presenting the results of the study. An example of a rationality-implyingVNM axiom is *transitivity*: if I prefer A over B, and B over C, then it is necessarily implied that I must (under conditions of rationality) also prefer A over C.

These axioms are considered particularly important in the formal decision sciences, although they are not always adequate to explain or justify human reasoning or motivation, as an ample body of literature in behavioral economics has shown.

Because this is a pilot study, we eschew the use of statistical and quantitative measures of performance, such as precision and accuracy, that these models are traditionally subjected to. In the last section of this paper, we comment on how these quantitative measures could potentially be applied, and the issues that may arise in practice, in a larger-scale, more controlled version of this preliminary work.

2. **Pilot Study**

*2.1 Spatial Reasoning*

We construct and apply spatial reasoning prompts, along the lines discussed earlier, to the DALL-E 2 model. Below, the 10 pairs of prompts we used are enumerated, along with the DALL-E output (four images per prompt). The experiment was conducted in December 2022 using a DALL-E browser-based interface that OpenAI had made available. The correct outputs, as graded by us, are identified using a green rectangle box. The remainder were deemed to be either incorrect or, at best, ambiguous.

| Prompt 1 / Prompt 2 | DALL-E Output |
|---|---|
| an apple in <u>front</u> of an orange / an apple <u>behind</u> an orange | 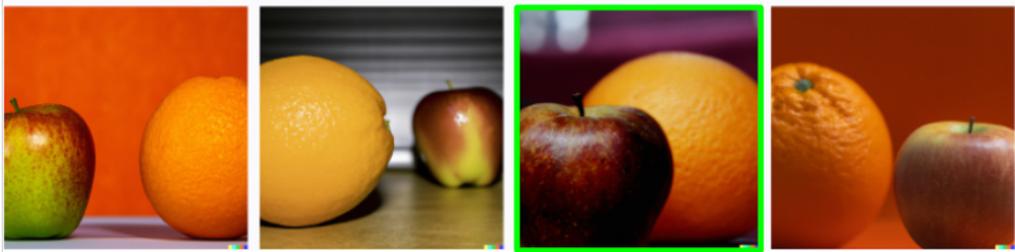 |
| an apple on the <u>left</u> of an orange / an apple on the <u>right</u> of an orange | 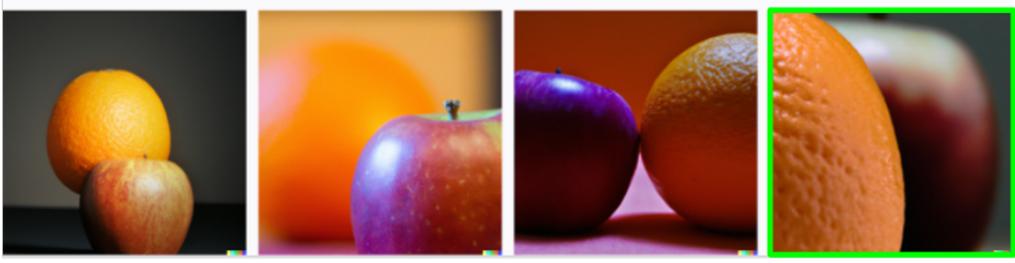 |
| | 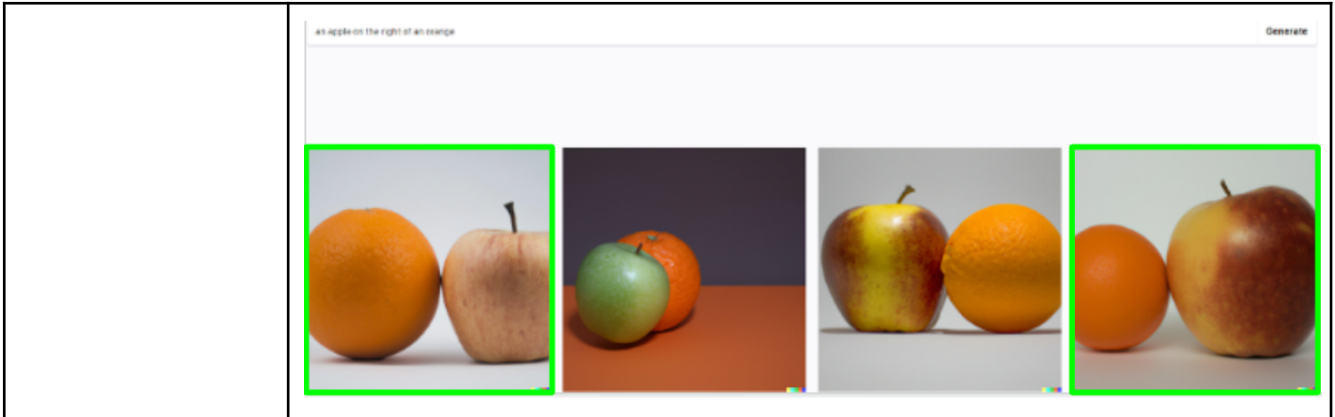 |
| an apple on the <u>bottom</u> of an orange /<br>an apple on <u>top</u> of an orange | 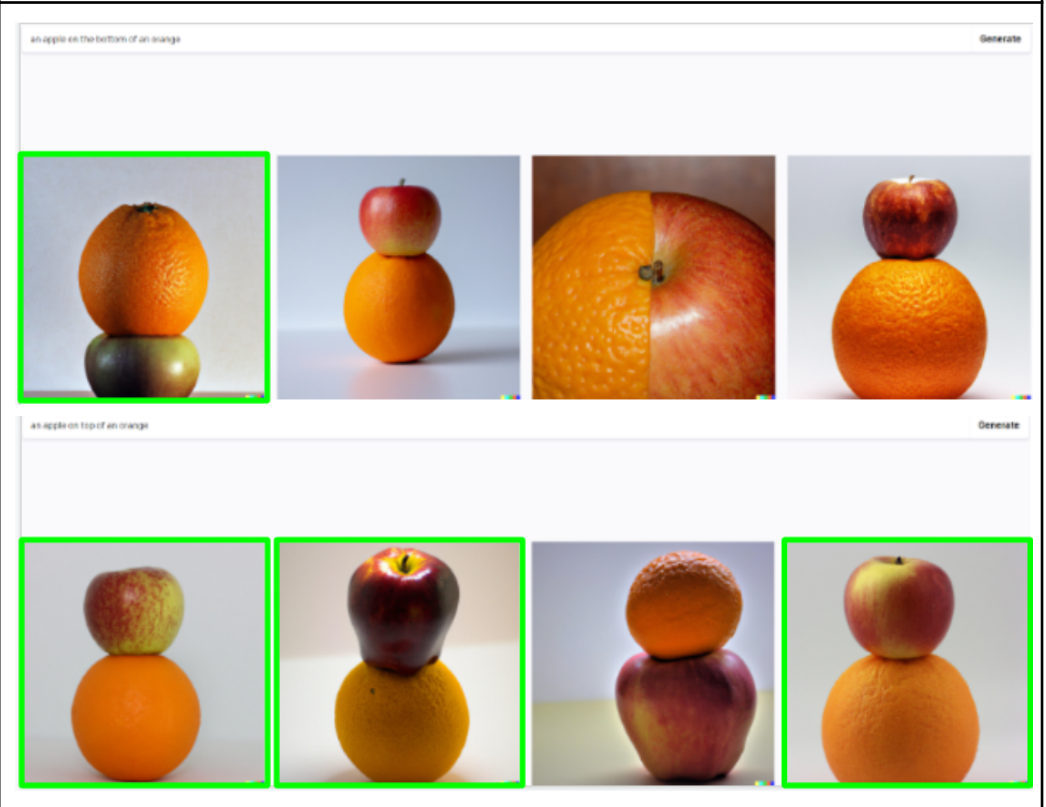 |
| a black <u>square</u> inside of a red <u>circle</u> /<br>a black <u>circle</u> inside of a red <u>square</u> | 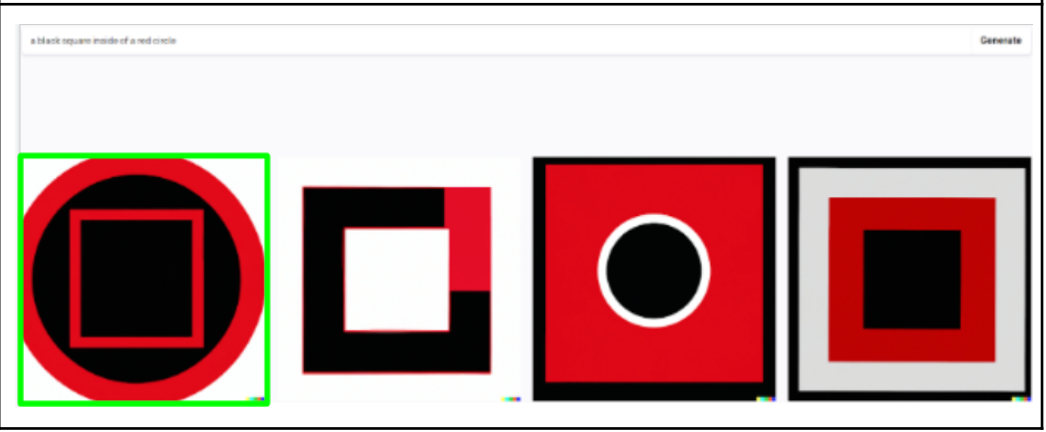 |
| | 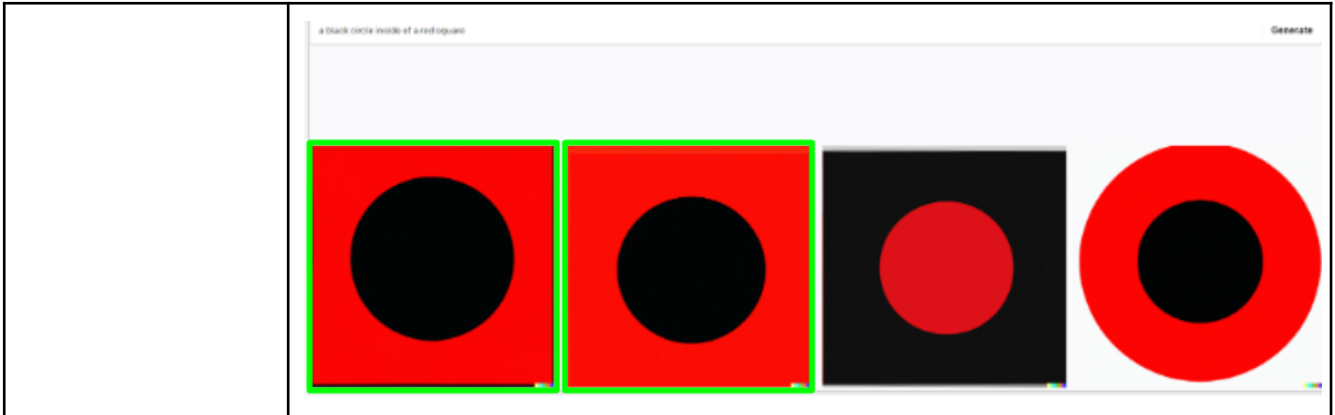 |
| a circle inside of a square / a square inside of a circle | 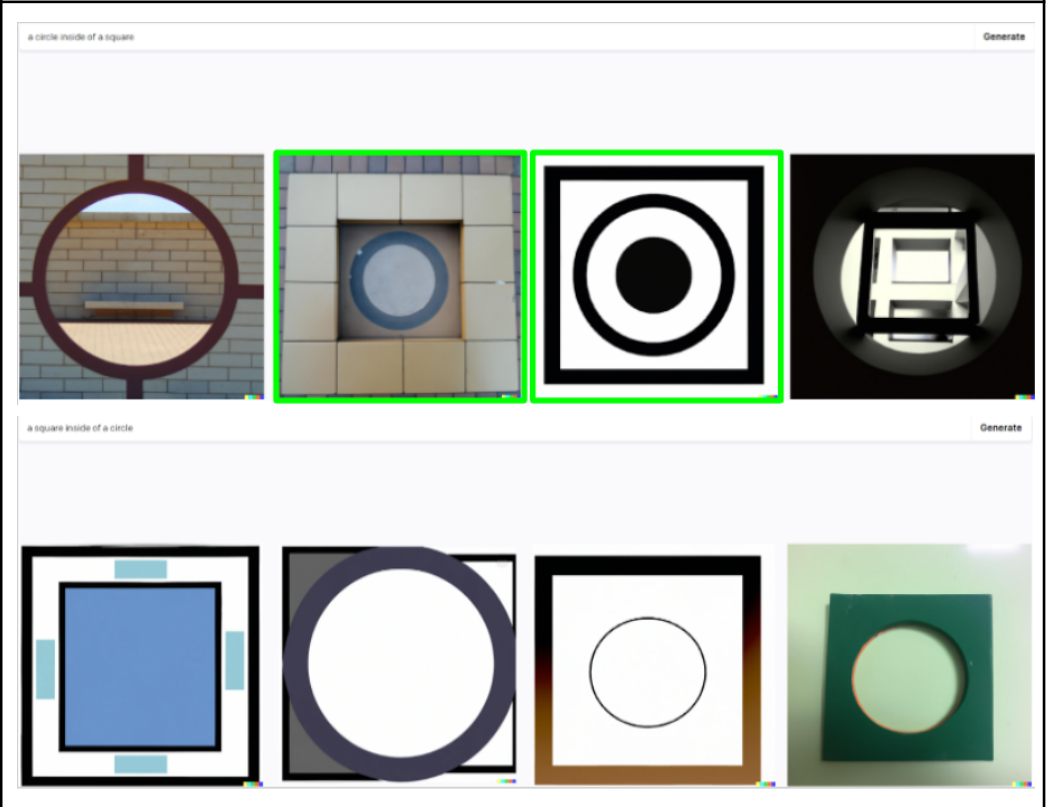 |
| top is black and bottom is red / top is red and bottom is black | 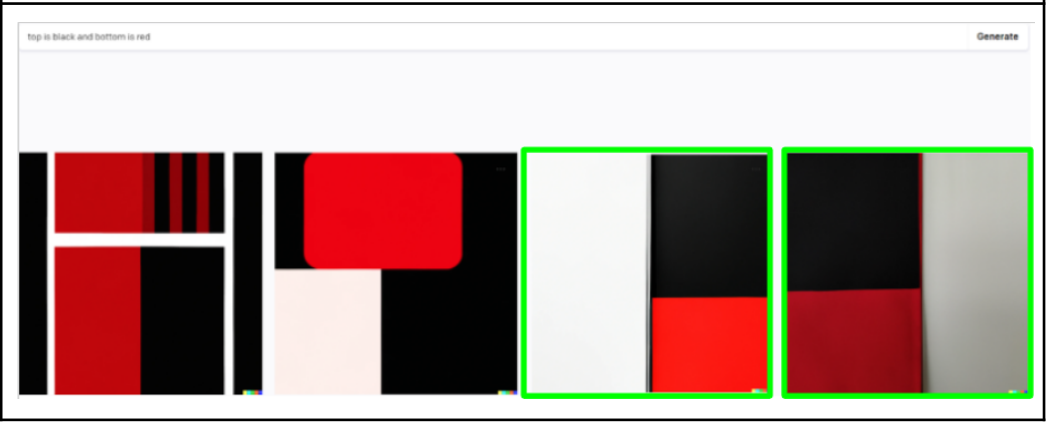 |
| | 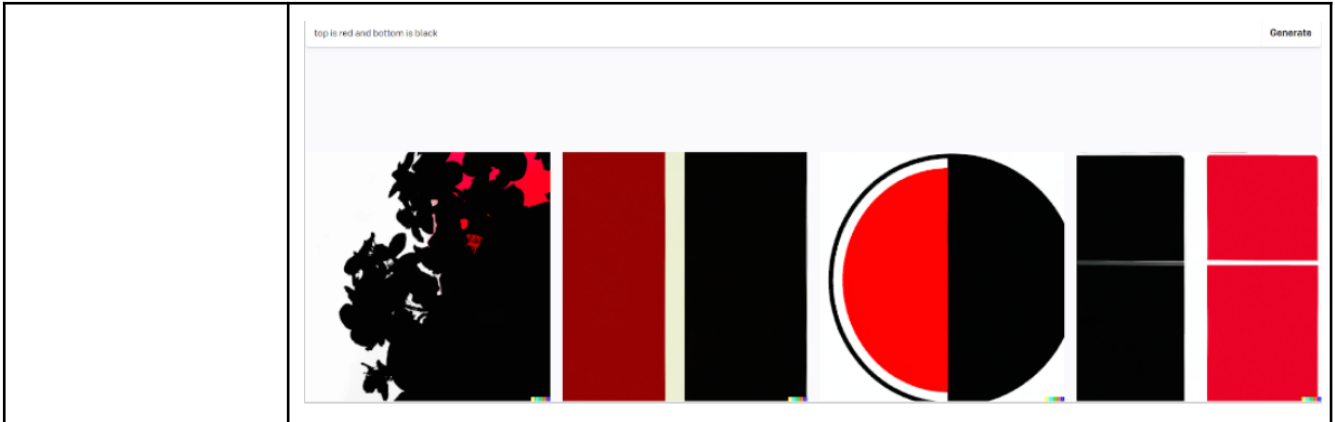 |
| left side is black and right side is red / left side is red and right side is black | 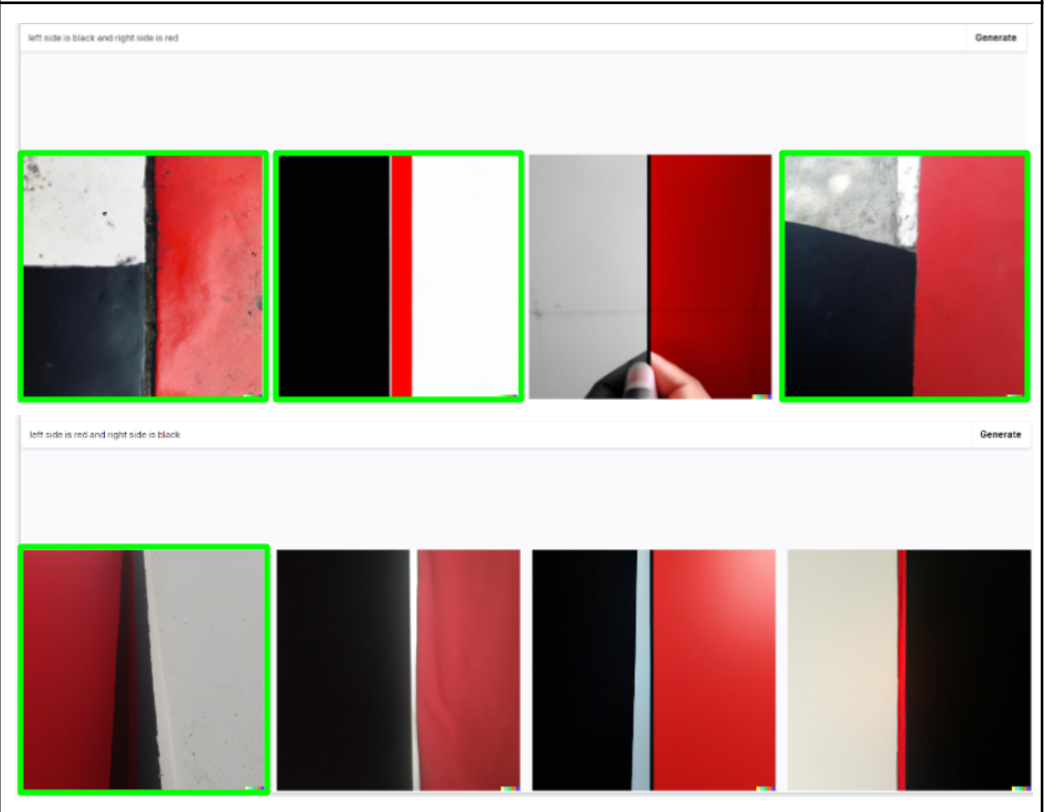 |
| a person standing right in front of the Eiffel Tower / a person standing 5 miles from the Eiffel Tower | 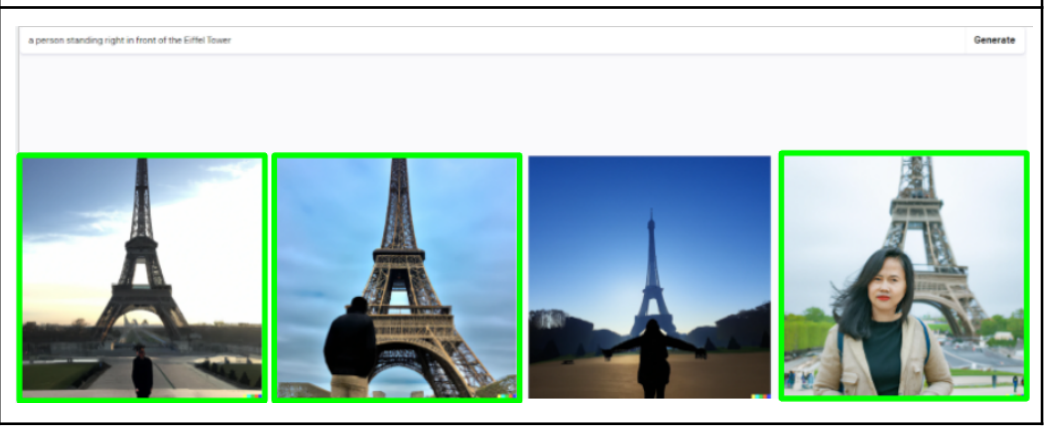 |
| | 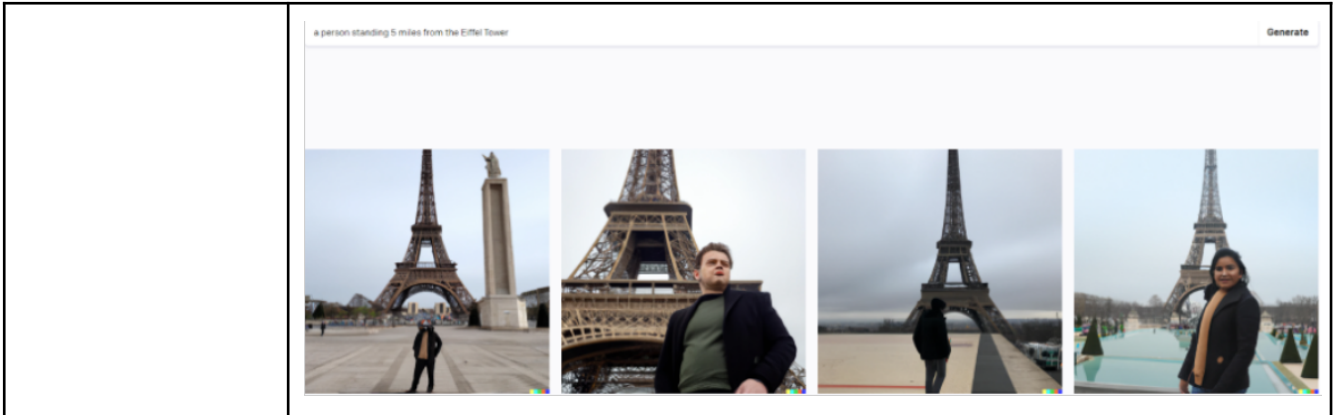 |
| a person with his <u>face</u> towards a car / a person with his <u>back</u> towards a car | 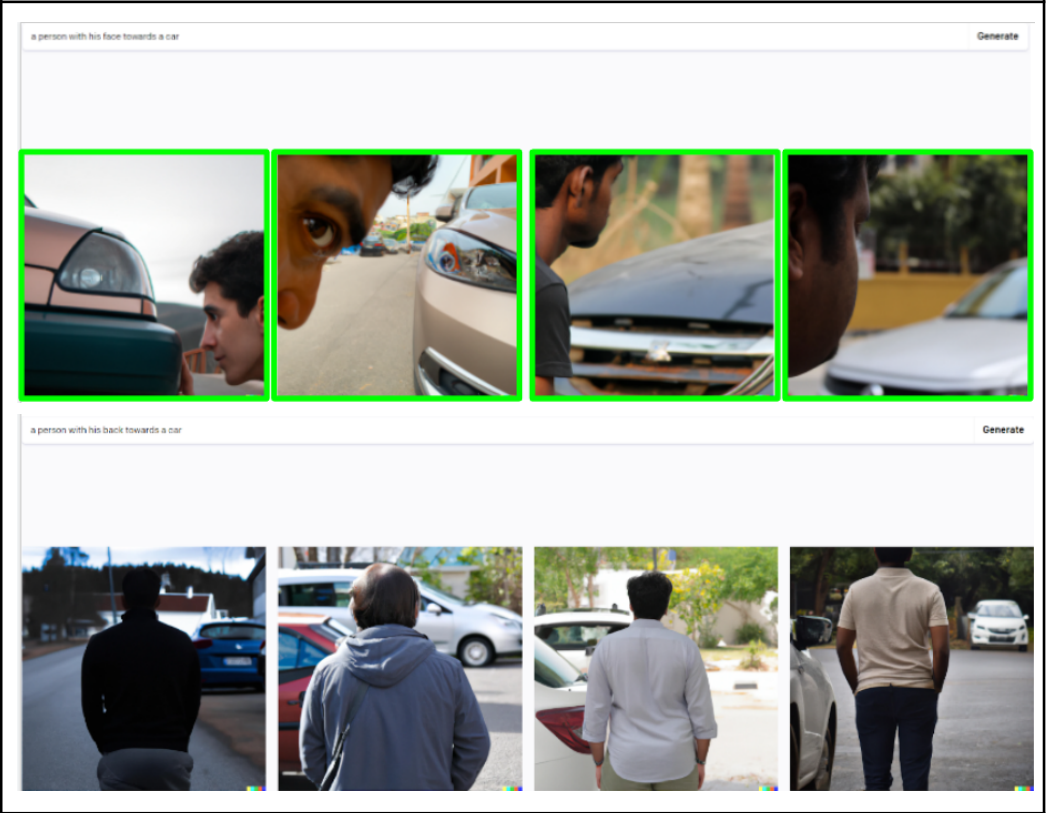 |
| a person with his hand pointing <u>towards</u> a car / a person with his hand pointing <u>away</u> from a car | 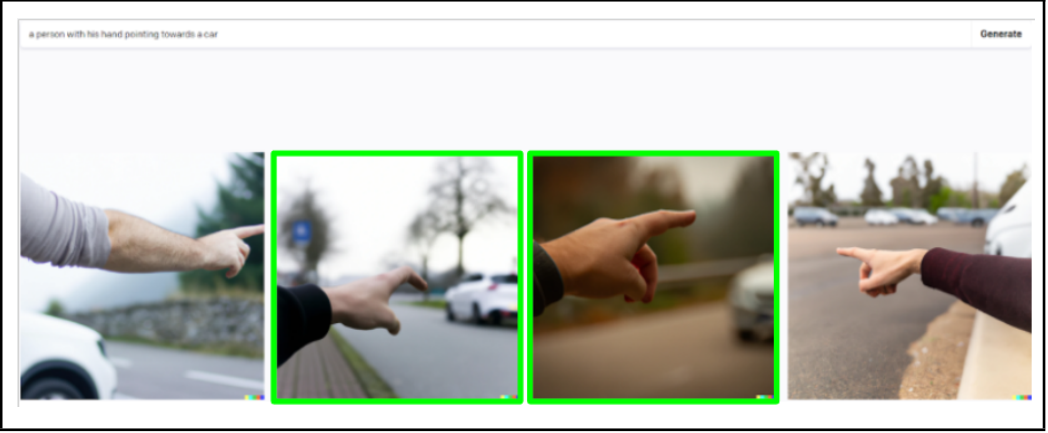 |

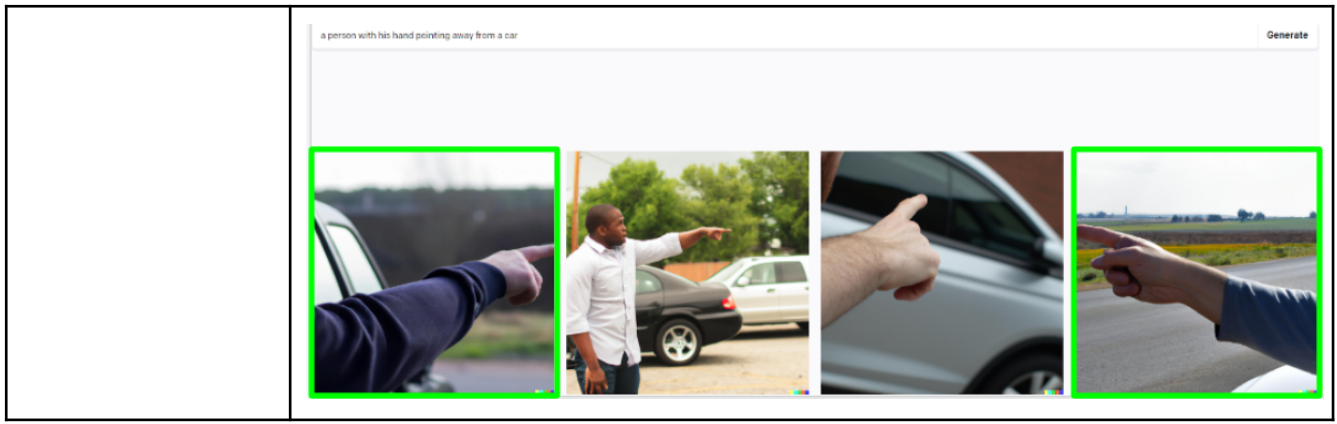

In qualitatively analyzing the results, we find that DALL-E 2 is generally able to generate at least one correct image per prompt, and that in six out of the ten prompt pairs it is able to generate a correct image for each of the prompts in the pair. An interesting range of 'extreme' behavior is also observed across some prompt-pairs. For example, in response to the prompt 'a person with his <u>face</u> towards a car' the model produces images that are all reasonable; however, in response to its counterpart prompt 'a person with his <u>back</u> towards a car' its outputs are more ambiguous and based on an interpretation of the language that we did not intend (but that others might argue are reasonable, showing the importance of having multiple independent post-hoc annotations in a more complete study). Tellingly, there is no pair of prompts where all generated images are correct for both prompts.

## 2.2 Decision-making

Broadly defined, rational decision-making has been theorized to involve a number of different axioms, especially under the seminal Von Neumann-Morgenstern utility theorem. A full evaluation of a model like ChatGPT on these axioms is beyond the scope of this pilot study. Instead, we evaluate the model on two axioms: the independence axiom and the transitivity axiom. We also evaluate other (non-VNM related) decision-making abilities by presenting to the model, bets that are conditioned on random events, bets with asymmetric outcomes, decisions encapsulating Savage's *Sure Thing* principle, and other complex bet structures like nested bets. Unlike other work that has sought to evaluate ChatGPT in an adversarial fashion, we construct our problems in a simple and uniform language without any intent of 'tricking' it. All prompts and responses are presented as is; in other words, prompts are not adjusted intentionally or dynamically to obtain a mal-response. Each experiment is conducted in a single, short session which usually comprises a few prompts presented in sequence. Each subsection below is relatively self-contained.

**2.2.1 Evaluating ChatGPT on the VNM independence axiom**

In the *Background*, we mentioned *transitivity* as an important VNM-rationality axiom. Another such axiom is *independence,* which concerns 'compound' lotteries[4] and can be formally expressed as saying that, for any lottery N and real-valued p ∈ (0,1],

$$L \preccurlyeq M \; iff \; pL + (1 - p)N \preccurlyeq pM + (1 - p)N$$

Here, L and M are also lotteries. As the equation suggests, if I prefer M over L (independent of another lottery N), then I should continue to prefer M over L if both are combined in the same way with any other lottery N (thereby yielding a compound lottery). To test ChatGPT for independence, we first state our preference regarding two items, then construct two different bet questions with equally likely outcomes that differ only in one of the two outcomes of each bet. We ask ChatGPT about which one we should choose.

---

[4] A lottery here may be described as a probabilistic bet, with a well-defined expected value; it can also colloquially be thought of as a 'decision' with outcomes decided under conditions of uncertainty and with an expected payoff: what the individual decides is often based on their risk-tolerance, but under VNM utility theory, some decisions are not rational *regardless* of the individual's risk-tolerance.

| Prompts | ChatGPT answers | Comment |
|---|---|---|
| P1: if i prefer watch over egg. if I am given a bet that has 50% chance to win a watch and 50% chance to win a computer and another bet that has 50% chance to win a egg and 50% chance to win a computer, which bet should I take? | ChatGPT1: Based on your preference, you should take the bet that has a 50% chance of winning a watch and 50% chance of winning a computer. This bet aligns with your preference of a watch over an egg and also offers the chance of winning a computer, which could be seen as a bonus. | ChatGPT shows a reasoning process consistent with the independence axiom and comes to the right (rational) conclusion. Even when we indicate our preference to be contrary to the commonsensical ones, ChatGPT can still reason using this fake preference and return a rational decision. |
| P2: if i prefer egg over watch. if I am given a bet that has 50% chance to win a watch and 50% chance to win a computer and another bet that has 50% chance to win a egg and 50% chance to win a computer, which bet should I take? | ChatGPT2: Based on your preference, you should take the bet that has a 50% chance of winning an egg and 50% chance of winning a computer. This bet aligns with your preference of an egg over a watch and also offers the chance of winning a computer, which could be seen as a bonus. | |

The result above suggests that ChatGPT's responses are aligned somewhat with the independence axiom. As a robustness check, even when we indicate our preference to be contrary to the commonsensical ones, ChatGPT can still reason using this fake preference. Furthermore, ChatGPT seems to understand from the context that we want to maximize our expected gain, even when we do not say so explicitly. An interesting question to explore for future research is whether we can get it to *minimize* expected gain when we instruct it to do so, similar to the contrastive prompting we used for testing DALL-E 2 on spatial reasoning.

### 2.2.2 Evaluating ChatGPT on the VNM transitivity axiom

To test ChatGPT on VNM *transitivity*, we construct three bets, all of which have equally likely but different outcomes. We state our preference among the three bets. Our preference always favors one bet over another bet and favors the other bet over yet another bet. Hence, we have a clear preference for the first bet over the last one. Then, we ask ChatGPT for our preference involving the first and last bet. The first prompt may contain preferences that violate common sense, such as preferring an egg over a TV. The preference in the second prompt aligns with commonsense.

| Prompts | ChatGPT answers | Comment |
|---|---|---|
| P1: if i prefer bet A that has 50% chance to win a watch and 50% chance to win a computer over bet B that has 50% chance to win a egg and 50% chance to win a computer, and i prefer bet B over bet C that has 50% chance to win a tv and 50% chance to win a computer. What is my preference involving bet A and bet C? | ChatGPT1: Based on your preferences, we can conclude that you prefer the option of winning a computer over winning any other item in all three bets. So, it can be said that you don't have a preference between bet A and bet C as both offer a 50% chance of winning a computer. | Both interactions show reasoning and conclusions that violate the transitivity axiom. |
| P2: if i prefer bet A that has 50% chance to win a car and 50% chance to win a computer over bet B that has 50% chance to win a tv and 50% chance to win a computer, and i prefer bet B over bet C that has 50% chance to win a egg and 50% chance to win a computer. What is my preference involving bet A and bet C? | ChatGPT2: Based on your preferences, it can be concluded that you prefer the option of winning a car in bet A over winning a TV in bet B and an egg in bet C. However, as both bet A and bet C offer a 50% chance of winning a computer, it can be said that you do not have a preference between bet A and bet C. | |

Unlike the test with the independence axiom, the result above suggests that the transitivity axiom is more challenging for ChatGPT. In both prompts, we state our preference for bet A over bet B, and for bet B over bet C. Following the transitivity axiom, ChatGPT should reasonably infer that we prefer bet A over bet C. However, not only does it get the wrong answer (that we do not have a preference for bet A over bet C) in both cases but its reasoning is also incorrect. In the first answer, ChatGPT says that 'you prefer the option of winning a computer over winning any other item in all three bets' which seems superficially correct, but is specious (and incorrect). What should be inferred from the first prompt is that we prefer winning a watch over winning an egg, and over winning a TV. Given the second prompt, ChatGPT *does* identify this preference relationship between the three items. However, it again mentions computers, which is irrelevant in the context of this problem.

### 2.2.3 Conditioning Bets on Random Events

In the first two experiments, we state the probabilities of possible outcomes explicitly. In the next experiment, we condition the probabilities on the result of a random event, such as the outcome of a coin flip. The reason is to control for the possibility that ChatGPT may simply not be equipped to handle 'explicit' probability estimates robustly. To test whether ChatGPT can choose the outcome with the highest expected gain, we start with the simplest formulation. We first confirm that ChatGPT's preference for an egg and a watch is aligned with those items' (commonsense) economic value. This is the basis for our further evaluation as it would allow us to treat the preference as given when evaluating rationality. Then, we construct a bet question with equally likely outcomes and a net positive expected gain. We ask ChatGPT which outcome we should bet on to maximize the expected gain. We ask once for each possible outcome.

| Prompts | ChatGPT answers | Comment |
| --- | --- | --- |
| P1: does a watch usually cost more than an egg? | ChatGPT1: The cost of a watch can vary greatly depending on many factors such as the brand, materials used, and features it has. Some watches can cost just a few dollars while others can cost tens of thousands of dollars. On the other hand, the average cost of an egg is relatively low, often just a few cents. So, it's difficult to make a general comparison between the cost of a watch and an egg. However, it's safe to say that in most cases, a watch will cost more than an egg. | ChatGPT shares the same preference for a watch and an egg as an average human. However, for all three prompts, ChatGPT's reasoning seems to be incorrect and it made the wrong (irrational) decision. |
| P2: A fair coin is flipped. If the coin comes up heads, then I win a watch. If it comes up tails, then I lose a egg. should I bet on heads to maximize my expected gains? | ChatGPT2: In this scenario, betting on heads would not be a wise decision to maximize your expected gains, as the potential rewards and losses are not equal. The expected gain from betting on heads is 0, as the probability of winning a watch (which is worth more than an egg) is exactly the same as the probability of losing an egg. On average, you would break even if you bet on heads every time.<br><br>It's important to consider the potential outcomes and their corresponding probabilities when making decisions, in order to make informed choices that maximize your expected gains. In this case, betting on heads would not result in any expected gain, so it would not be a good choice. | |
| P3: A fair coin is flipped. If the coin comes up heads, then I win a watch. If it comes up tails, then I lose a egg. | ChatGPT3: No, betting on tails would not be a wise decision to maximize your expected gains either. The expected gain from betting on tails would also be 0, as the probability of losing a watch (which is | |
| should I bet on tails to maximize my expected gains? | worth more than an egg) is exactly the same as the probability of winning an egg. On average, you would break even if you bet on tails every time.<br><br>In this scenario, there is no way to maximize your expected gains, as the rewards and losses are not equal and the coin flip is fair. You would have an equal probability of winning a watch or losing an egg, and on average, you would break even over many coin flips. It's best to not make a bet in this case, as you would simply be exchanging one item for another with no expected change in your overall value. | |
| P4: A fair coin is flipped. If the coin comes up heads, then I win a watch. If it comes up tails, then I lose a egg. should I not bet to maximize my expected gains? | ChatGPT4: Yes, in this scenario, not making a bet would be the best choice to maximize your expected gains. Since the potential rewards and losses are not equal (a watch is worth more than an egg), betting on either heads or tails would result in an expected loss. On average, you would break even if you made either bet every time, but you would not have an expected gain.<br><br>In this case, not making a bet would allow you to avoid any potential losses and maintain your current possessions (neither a watch nor an egg). Not betting would be the best way to maximize your expected gains, as you would avoid any exchange that could result in a net loss. | |

The results above suggest that ChatGPT's reasoning and decision are both incorrect. In the second response, ChatGPT responded that the potential rewards and losses are not equal, which is correct. However, in the next sentence, it says 'the expected gain from betting on heads is 0, as the probability of winning a watch (which is worth more than an egg) is exactly the same as the probability of losing an egg.' This is incorrect because even though the probability of winning a watch is the same as the probability of losing an egg, the expected gain from betting on heads is not zero. In the next two responses, ChatGPT continues to make such mistakes which leads to incorrect decisions.

In the previous experiment, prompts were only concerned with one bet. To test instead whether ChatGPT can identify the bet (or 'lottery') with a higher expected gain when presented with two bets, we construct two bets with equally likely outcomes. One of the bets has a positive net expected gain while the other has a negative net expected gain. In principle, this is a relatively easy decision. We ask ChatGPT about which bet we should choose. In the first prompt, abstract items are involved. In the second prompt, actual monetary value is used. We asked both versions, as there is some chance the model is better able to reason probabilistically when given one over the other.

| Prompts | ChatGPT answers | Comment |
|---|---|---|
| P1: if I am given a bet that If a coin comes up heads, then I win a watch. If it comes up tails, then I lose a egg. And another bet that If a coin comes up heads, then I win a egg. If it comes up tails, then I lose a watch. which bet should I take? | ChatGPT1: In this scenario, both bets have the same expected value. The expected value is the sum of the product of the outcome and its corresponding probability. For both bets, the probability of heads and tails are equal (1/2), and the outcomes of winning a watch or an egg are also equal in value.<br><br>Therefore, both bets have the same expected value, and it doesn't matter which bet you take. On average, you would break even over many coin flips. It's important to note that expected value is just an average and does not guarantee a particular outcome in each individual flip of the coin. It is possible to win a watch in one bet and lose a watch in another, but on average over many coin flips, you would break even. | In the first response, ChatGPT gives the false statement that the two bets have the same expected value and comes to the wrong conclusion. In the second response, ChatGPT correctly identified the correct choice with the correct reasoning process. |
| P2: if I am given a bet that If a coin comes up heads, then I win a $100. If it comes up tails, then I lose a $1. And another bet that If a coin comes up heads, then I win a $1. If it comes up tails, then I lose a $100. which bet should I take? | ChatGPT2: In this scenario, it would be best to take the first bet (win $100 and lose $1). This bet has a higher expected value compared to the second bet (win $1 and lose $100).<br><br>The expected value of the first bet is (1/2) * $100 + (1/2) * -$1 = $49.50, which is positive and represents the average amount you would expect to win over many coin flips.<br><br>The expected value of the second bet is (1/2) * $1 + (1/2) * -$100 = -$49.50, which is negative and represents the average amount you would expect to lose over many coin flips.<br><br>Therefore, taking the first bet would have a higher expected value, and would give you a better chance of making a profit in the long run compared to the second bet. | |

The results above suggest, interestingly, that ChatGPT is able to handle this type of selection problem better when it is given monetary values explicitly, rather than pairs of items where the values have to be inferred. Nevertheless, as had already been found for one of the previous prompts, ChatGPT has a preference for a watch over an egg. Rationally, it should also have selected the positive expectancy bet for the first prompt above.

In the previous prompt, we implicitly assumed that we prefer a watch over an egg, as an average person would have. What if we state our preference explicitly? In this experiment, we do so by stating our preference for a watch over an egg. Then, we pose to ChatGPT the same prompt as in the previous experiment, which it got wrong. In the second prompt, we switch our preference and ask the same question again.

| Prompt | ChatGPT answer | Comment |
| --- | --- | --- |
| P1: if i prefer watch over egg. if I am given a bet that If a coin comes up heads, then I win a watch. If it comes up tails, then I lose a egg. And another bet that If a coin comes up heads, then I win a egg. If it comes up tails, then I lose a watch. which bet should I take? | ChatGPT1: If you prefer a watch over an egg, you should take the first bet, where if a coin comes up heads, you win a watch and if it comes up tails, you lose an egg.<br><br>In this case, if the coin comes up heads, you get what you prefer (a watch) and if it comes up tails, you lose something that you value less (an egg). So, this bet gives you the best chance of getting what you want. | ChatGPT shows the correct reasoning process and gives the correct conclusion for both prompts, even when we switch our preferences. |
| P2: if i prefer egg over watch. if I am given a bet that If a coin comes up heads, then I win a watch. If it comes up tails, then I lose a egg. And another bet that If a coin comes up heads, then I win a egg. If it comes up tails, then I lose a watch. which bet should I take? | ChatGPT2: If you prefer an egg over a watch, then you should take the second bet, where if a coin comes up heads, you win an egg and if it comes up tails, you lose a watch.<br><br>In this case, if the coin comes up heads, you get what you prefer (an egg) and if it comes up tails, you lose something that you value less (a watch). So, this bet gives you the best chance of getting what you want. | |

Interestingly, stating the preference explicitly now leads to the model employing the correct reasoning process, and giving the correct answer for both prompts. The model would thus seem to be robust in some ways. This particular experiment would seem to confirm the importance of 'prompt engineering' in such experiments, even for models as linguistically capable as ChatGPT.

### 2.2.4 Bets with Asymmetric Probabilities

All previous experiments are concerned with bets with equally likely outcomes. To test ChatGPT on bets with outcomes that are not equally likely, we construct a bet question with asymmetric probabilities and then ask ChatGPT about how we should act to maximize expected gain.

| Prompts | ChatGPT answers | Comment |
| --- | --- | --- |
| P1: If I roll a die and it comes up with a number that is less than 2, then I get a watch. If it comes up with a number that is greater or equal to 2, then I give an egg? What should I do to maximize my expected gains? | ChatGPT1: Since a standard die has six sides, each side has an equal probability of showing up. So, the probability of rolling a number less than 2 is 1/6, and the probability of rolling a number greater than or equal to 2 is 5/6.<br><br>If you roll a number less than 2, you get a watch, which we can assume has a positive value for you. On the other hand, if you roll a number greater than or equal to 2, you must give away an egg, which we can assume has a negative value for you.<br><br>To maximize your expected gains, you should choose the option with the higher expected value. In this case, giving away an egg has a higher expected value because it happens 5 out of 6 times, so your expected loss would be 5/6 times the value of the egg. On the other hand, getting a watch has a lower expected value because it only happens 1 out of 6 times, so your expected gain would be 1/6 times the value of the watch.<br><br>Therefore, to maximize your expected gains, you should roll the die and, if the number is less than 2, take the watch, and if the number is greater or equal to 2, give away the egg. | ChatGPT can determine the correct probability for the bet. However, ChatGPT's claim that 'giving away an egg has a higher expected value' is wrong. Although part of its conclusion is ultimately correct, it is not complete. |

The response to this prompt is ambiguous. Even though ChatGPT can correctly reason about the asymmetric probability and part of ChatGPT's conclusion is correct (that we should roll the die to maximize the expected gain), it is derived from an incorrect intermediate step (that giving away an egg has a higher expected value and getting a watch has a lower expected value).

### 2.2.5 Savage's Sure Thing Principle

Beyond the VNM axioms, another important principle concerning decision-making is Savage's *Sure Thing* principle (first elucidated in 1954; see Pearl [29] for an accessible description), which states that [*Let f and g be any two lotteries*], *if a person prefers f to g, either knowing that the event B obtained, or knowing that the event not-B obtained, then he should prefer f to g even if he knows nothing about B.* To test ChatGPT for the Sure Thing principle, we first state our preference for two bets. Each of the two bets contains the same likelihood of winning an identical item (30% for the egg in the first prompt). Then we ask ChatGPT for our preference over another two bets. These two bets differ from the original bets only in the aforementioned item. By the Sure Thing principle, our preferences should not change. The second prompt changes abstract items to actual values.

| Prompt | ChatGPT answer | Comment |
| --- | --- | --- |
| P1: if I prefer bet A which has 60% chance to win a car, 10% chance to win a coffee, and 30% chance to win an egg to bet B which has 60% chance to win a computer, 10% chance to win nothing, and 30% chance to win an egg. I was given bet C which has 60% chance to win a car, 10% chance to win a coffee, and 30% chance to win a diamond. And another bet D which has 60% chance to win a computer, 10% chance to win nothing, and 30% chance to win a diamond. what is my preference involving bet C and bet D? | ChatGPT1: If you prefer bet A to bet B, it means that you value the prospects of winning a car and a coffee more highly than the prospects of winning a computer. In that case, your preference between bet C and bet D would depend on whether you value diamonds more highly than either cars or computers.<br><br>If you value diamonds more highly than cars, then you would prefer bet C, as it gives you a higher chance of winning a diamond. Similarly, if you value diamonds more highly than computers, then you would prefer bet D.<br><br>However, if you value diamonds less highly than both cars and computers, then you would have to make a decision based on which of the two you value more highly. If you value cars more highly than computers, you would prefer bet C, and if you value computers more highly than cars, you would prefer bet D. | ChatGPT gives incorrect reasoning and conclusion. For the second prompt, its application of the Sure Thing principle is more ambiguous. |
| P2: if I prefer bet A which has 60% chance to win $1000, 10% chance to win $0, and 30% chance to win $0 to bet B which has 60% chance to win $500, 10% chance to win $300, and 30% chance to win $0. I was given bet C which has 60% chance to win $1000, 10% chance to win $0, and 30% chance to win $1000. And another bet D which has 60% chance to win $500, 10% chance to win $300, and 30% chance to win $1000. what is my preference involving bet C and bet D? | ChatGPT2: If you prefer bet A to bet B, it means that you value the prospects of winning $1000 more highly than the prospects of winning a combination of $500 and $300.<br><br>In that case, your preference between bet C and bet D would depend on whether you value the chance of winning $1000 more highly than the combination of $500 and $300. If you value the chance of winning $1000 more highly, then you would prefer bet C, as it gives you a higher chance of winning $1000. Similarly, if you value the chance of winning $1000 more highly, then you would prefer bet D.<br><br>However, if you value the combination of $500 and $300 more highly than the chance of winning $1000, then you would have to make a decision based on which of the two you value more highly. If you value bet C more highly, you would prefer bet D, and if you value bet D more highly, you would prefer bet C. | |

ChatGPT's response to the first prompt does not seem to follow the Sure Thing principle, giving both incorrect reasoning and an incorrect conclusion. The response to the second prompt is more ambiguous. In its first response sentence, ChatGPT gives a correct statement that we prefer winning $1000 more than winning a combination of $500 and $300. In the middle of the

response, it is unclear which $1000 ChatGPT refers to. In the last sentence, ChatGPT gives an incorrect conclusion that if we value bet C more highly, we would prefer bet D and vice versa.

**2.2.6 Other Complex Betting Structures**

Finally, we note that the bets we have presented thus far follow a very simple and relatively symmetric structure. In decision science, it is possible to construct increasingly complicated bet questions, such as a pair of bets with different likelihoods of outcomes, and bets with another bet as one of their outcomes (nested bets). To test ChatGPT on such bets, we construct two prompts: the first prompt contains two bets with three outcomes with different likelihoods of occurring. The second prompt contains a nested bet.

| Prompt | ChatGPT answer | Comment |
|---|---|---|
| P1: if I am given a bet that has 50% chance to win $1000, 30% chance to lose $500, and 20% chance to lose $100 and another bet that has 50% chance to win $2000, 40% chance to lose $500, and 10% chance to win $300, which bet should I take? | ChatGPT1: To calculate the expected value of each bet, you can multiply the amount of each outcome by its corresponding probability, and then sum the results.<br>For the first bet:<br>● 50% chance to win $1000, which has an expected value of 0.5 * $1000 = $500<br>● 30% chance to lose $500, which has an expected value of -0.3 * $500 = -$150<br>● 20% chance to lose $100, which has an expected value of -0.2 * $100 = -$20<br>So the expected value of the first bet is $500 - $150 - $20 = $330.<br>For the second bet:<br>● 50% chance to win $2000, which has an expected value of 0.5 * $2000 = $1000<br>● 40% chance to lose $500, which has an expected value of -0.4 * $500 = -$200<br>● 10% chance to win $300, which has an expected value of 0.1 * $300 = $30<br>So the expected value of the second bet is $1000 - $200 + $30 = $830.<br>Based on these calculations, the second bet has a higher expected value, so it might be a better choice if you're willing to take the risk. However, it's important to remember that expected value does not guarantee a profit, and there is always a risk involved when gambling. You should always be aware of the risks and be prepared for the possibility of losing, no matter how small the chance. | ChatGPT outputs the correct reasoning process and correct conclusion for both prompts. |
| \<br><br>P2: if I am given a bet that has 50% chance to win $1000 and 50% chance to enter another bet which has 50% chance to win | ChatGPT2: To calculate the expected value of each bet, you can use the concept of conditional probability.<br>For the first bet: | |
| $100, and 50% chance to lose $100, and another bet that has 50% chance to win $1000, 50% chance to lose $300, which bet should I take? | • If you win the first bet, you'll enter another bet with 50% chance to win $100 and 50% chance to lose $100. The expected value of this second bet is 0, since the potential gains and losses cancel each other out. So the expected value of the first bet is 0.5 * 0 + 0.5 * $1000 = $500.<br>For the second bet:<br>• The expected value of the second bet is 0.5 * $1000 - 0.5 * $300 = $350.<br>Based on these calculations, the first bet has a higher expected value, so it might be a better choice if you're willing to take the risk. However, it's important to remember that expected value does not guarantee a profit, and there is always a risk involved when gambling. You should always be aware of the risks and be prepared for the possibility of losing, no matter how small the chance. | |

Given the previous results, we would have expected the model to have completely failed (or yield generic results that did not answer the specific question); however, we find that the model outputs both the correct reasoning process and the correct answers for both prompts. This finding is somewhat, but not completely, surprising. It also highlights the need to conduct a series of neutral prompt-based evaluations when testing the psycholinguistic and cognitive properties of such black-box models. A wrong answer in response to a seemingly simpler prompt does not *imply* a wrong answer in (what we may assume to be) a more complex prompt. The last instance also makes it problematic to argue that ChatGPT got the right answer for the 'wrong' reasons as the reasoning in that instance seems rigorous and would have been accepted as such if a human had given it.

## Discussion

We conclude with a note on scaling up these evaluations and quantifying the qualitative observations we stated earlier. We believe it is easier to provide more quantitative estimates for the DALL-E 2 spatial reasoning experiments as determining when the model is behaving properly seems more evident. In looking at the ten pairs of prompts we tried for the spatial reasoning test, if we were to attempt a quantitative analysis, at least two reasonable measures are suggested: (i) non-strict[5] completeness (recall): for how many prompts (a similar exercise can be done at the level of prompt-pairs) has at least one correct image been generated?; (ii) precision: for how many prompts have all images been correctly generated?; (iii) accuracy: What is the ratio of correct (to all) images generated per prompt? For the last measure in particular, micro- and macro-based averaging can both be used, but the simplest way to compute it is just to count the number of 'green boxes' in the pilot study results for DALL-E 2 and divide it by the total number of images generated. This simple version of accuracy, according to our results, would then be 32/40=80% for the study conducted, while the precision is 1/20=5% and non-strict completeness is 16/20=80%. If the first two measures are computed at the level of prompt pairs, precision is 0% while recall is 6/10=60%. While these are not enough data points to draw a robust conclusion, they confirm our intuition that these models may be less sensitive than we are initially led to believe from ad hoc case studies of their proficiency.

One issue with using traditional AI benchmarking methods (such as the use of tests that have an 'answer key') or their close cousin in psychology tests of a more subjective nature (e.g., finite interpretations of answer sets in personality tests such as [27]) is that generative models, by definition, produce outputs that may not be predicted, and hence 'graded', in advance. There is also some evidence to show that, in closed-set evaluations of the traditional kind, large language models may end up getting the right answers for the 'wrong' reasons [24]. We saw some evidence of this in our ChatGPT experiments, but not always. In some fairly complex cases, it was surprisingly able to draw the right conclusion and output a correct reasoning process. Ultimately, only a rigorous and empirical methodology, applied over a sufficiently large set of prompts constructed in a neutral manner can settle such issues without being unduly biased against the model or its ability to learn. Therefore, the evaluation of such models remains an interesting problem that needs to be taken a careful look at, even as newer and more complex generative models keep subsuming the earlier ones (both in academic research and public opinion) with each passing year.

---

[5] Non-strict here is taken to mean that at least one correct item was generated. Since hypothetically, the total number of possible correct images that could be generated for a given prompt is (usually) infinite in an open-ended space of possible generations, a true recall number (which depends on the number of true positives and false negatives in an underlying 'ground truth') cannot be estimated.

At the same time, the line is getting increasingly blurred between *task* benchmarking and *cognitive* benchmarking, of the kind that was proposed in this paper (but also in others). A recent work, for instance, conducted an evaluation of ChatGPT on 'reasoning, hallucination, and interactivity' [28]. However, they still used discriminatively constructed benchmarks, which may be a source of potential bias, as noted above. In other research we have conducted on evaluating these models on commonsense categories, we found that generative models can sometimes produce reasonable answers that are not in the original set of answers provided to them. Other work has shown that much can depend on how the prompt is administered to the model. While in the ideal world, a model would be able to address all of these problems, we also cannot ignore the potential limitations of such restrictive testing of models that were designed to be generative.

# References

[1] Vaswani, A., Shazeer, N., Parmar, N., Uszkoreit, J., Jones, L., Gomez, A. N., ... & Polosukhin, I. (2017). Attention is all you need. *Advances in neural information processing systems*, *30*.

[2] Devlin, J., Chang, M. W., Lee, K., & Toutanova, K. (2018). Bert: Pre-training of deep bidirectional transformers for language understanding. *arXiv preprint arXiv:1810.04805*.

[3] Liu, Y., Ott, M., Goyal, N., Du, J., Joshi, M., Chen, D., ... & Stoyanov, V. (2019). Roberta: A robustly optimized bert pretraining approach. *arXiv preprint arXiv:1907.11692*.

[4] Bianchi, F., Kalluri, P., Durmus, E., Ladhak, F., Cheng, M., Nozza, D., ... & Caliskan, A. (2022). Easily accessible text-to-image generation amplifies demographic stereotypes at large scale. *arXiv preprint arXiv:2211.03759*.

[5] Ettinger, A. (2020). What BERT is not: Lessons from a new suite of psycholinguistic diagnostics for language models. *Transactions of the Association for Computational Linguistics*, *8*, 34-48.

[6] Rogers, A., Kovaleva, O., & Rumshisky, A. (2021). A primer in BERTology: What we know about how BERT works. *Transactions of the Association for Computational Linguistics*, *8*, 842-866.

[7] Gessler, L., & Schneider, N. (2021). BERT has uncommon sense: Similarity ranking for word sense BERTology. *arXiv preprint arXiv:2109.09780*.

[8] Jing, K., & Xu, J. (2019). A survey on neural network language models. *arXiv preprint arXiv:1906.03591*.

[9] Maus, N., Chao, P., Wong, E., & Gardner, J. (2023). Adversarial Prompting for Black Box Foundation Models. *arXiv preprint arXiv:2302.04237*.

[10] Sun, L., Hashimoto, K., Yin, W., Asai, A., Li, J., Yu, P., & Xiong, C. (2020). Adv-bert: Bert is not robust on misspellings! generating nature adversarial samples on bert. *arXiv preprint arXiv:2003.04985*.

[11] Thorp, H. H. (2023). ChatGPT is fun, but not an author. *Science*, *379*(6630), 313-313.

[12] Microsoft Bets Billions on DALL-E and ChatGPT Maker OpenAI. URL: https://risnews.com/microsoft-bets-billions-dall-e-and-chatgpt-maker-openai

[13] What is generative AI? McKinsey and Company. URL: https://www.mckinsey.com/featured-insights/mckinsey-explainers/what-is-generative-ai

[14] A number of papers on large language models were presented (including some in their own dedicated session) in CogSci 2022: https://cognitivesciencesociety.org/cogsci-2022/

[15] Collins, K. M., Wong, C., Feng, J., Wei, M., & Tenenbaum, J. B. (2022). Structured, flexible, and robust: benchmarking and improving large language models towards more human-like behavior in out-of-distribution reasoning tasks. *arXiv preprint arXiv:2205.05718*.

[16] Kejriwal, M., & Tang, Z. (2022). Evaluating Language Representation Models on Approximately Rational Decision Making Problems. In *Proceedings of the Annual Meeting of the Cognitive Science Society* (Vol. 44, No. 44).

[17] Basner, M., Savitt, A., Moore, T. M., Port, A. M., McGuire, S., Ecker, A. J., ... & Gur, R. C. (2015). Development and validation of the cognition test battery for spaceflight. *Aerospace medicine and human performance*, *86*(11), 942-952.

[18] Shaw, R. C., & Schmelz, M. (2017). Cognitive test batteries in animal cognition research: evaluating the past, present and future of comparative psychometrics. *Animal cognition*, *20*(6), 1003-1018.

[19] Mathuranath, P. S., Nestor, P. J., Berrios, G. E., Rakowicz, W., & Hodges, J. R. (2000). A brief cognitive test battery to differentiate Alzheimer's disease and frontotemporal dementia. *Neurology*, *55*(11), 1613-1620.

[20] Kejriwal, M., Santos, H., Mulvehill, A. M., & McGuinness, D. L. (2022). Designing a strong test for measuring true common-sense reasoning. *Nature Machine Intelligence*, *4*(4), 318-322.

[21] Marcus, G., Davis, E., & Aaronson, S. (2022). A very preliminary analysis of dall-e 2. *arXiv preprint arXiv:2204.13807*.

[22] Leivada, E., Murphy, E., & Marcus, G. (2022). DALL-E 2 Fails to Reliably Capture Common Syntactic Processes. *arXiv preprint arXiv:2210.12889*.

[23] Borji, A. (2022). Generated faces in the wild: Quantitative comparison of stable diffusion, midjourney and dall-e 2. *arXiv preprint arXiv:2210.00586*.

[24] Shen, K., & Kejriwal, M. (2023). An experimental study measuring the generalization of fine‑tuned language representation models across commonsense reasoning benchmarks. *Expert Systems*, e13243.

[25] Tang, Z., & Kejriwal, M. (2022). Can Language Representation Models Think in Bets?. *arXiv preprint arXiv:2210.07519*.

[26] Von Neumann, J., & Morgenstern, O. (1944). Theory of games and economic behavior.

[27] Tett, R. P., & Christiansen, N. D. (2007). Personality tests at the crossroads: A response to Morgeson, Campion, Dipboye, Hollenbeck, Murphy, and Schmitt (2007). *Personnel psychology*, *60*(4), 967-993.

[28] Bang, Y., Cahyawijaya, S., Lee, N., Dai, W., Su, D., Wilie, B., ... & Fung, P. (2023). A Multitask, Multilingual, Multimodal Evaluation of ChatGPT on Reasoning, Hallucination, and Interactivity. *arXiv preprint arXiv:2302.04023*.

[29] Pearl, J. (2016). The sure-thing principle. Journal of Causal Inference, 4(1), 81-86.